\documentclass[letterpaper, 10 pt, conference]{ieeeconf}  

\IEEEoverridecommandlockouts                              

\usepackage{graphics} 
\usepackage{graphicx}
\usepackage{subcaption}
\usepackage{amsmath} 
\usepackage{amssymb}  
\usepackage{bm}

\usepackage[font=small,labelfont=bf]{caption}

\usepackage[colorlinks]{hyperref}

\title{\LARGE \bf
Online Photometric Calibration of Auto Exposure Video\\[2mm] for Realtime Visual Odometry and SLAM
}

\author{Paul Bergmann \\  {\tt\small bergmann@in.tum.de} \and Rui Wang \\  {\tt\small wangr@in.tum.de}  \and Daniel Cremers \\  {\tt\small cremers@tum.de} 
\thanks{All authors are with the Faculty of Computer Science, Technical University of Munich, Germany, Arcisstr. 21, 80333 Munich}}%

\usepackage{mathtools}

\begin{document}

\maketitle
\thispagestyle{empty}
\pagestyle{empty}

\begin{abstract}

Recent direct visual odometry and SLAM algorithms have demonstrated impressive levels of precision. However, they require a photometric camera calibration in order to achieve competitive results.  Hence, the respective algorithm cannot be directly applied to an off-the-shelf-camera or to a video sequence acquired with an unknown camera.  In this work we propose a method for online photometric calibration which enables to process auto exposure videos with visual odometry precisions that are on par with those of photometrically calibrated videos. Our algorithm recovers the exposure times of consecutive frames, the camera response function, and the attenuation factors of the sensor irradiance due to vignetting. Gain robust KLT feature tracks are used to obtain scene point correspondences as input to a nonlinear optimization framework. We show that our approach can reliably calibrate arbitrary video sequences by evaluating it on datasets for which full photometric ground truth is available. We further show that our calibration can improve the performance of a state-of-the-art direct visual odometry method that works solely on pixel intensities, calibrating for photometric parameters in an online fashion in realtime.

\begin{keywords}
Photometric calibration, online calibration, visual odometry, visual SLAM. 
\end{keywords}

\end{abstract}

\section{Introduction and Related Work}

Recently a number of direct methods for visual odomertry and visual SLAM such as DSO or LSD-SLAM
have been proposed, working only on pixel intensities  \cite{dso}\cite{lsd_slam}. They all rely on the underlying assumption, that a scene point appears with constant brightness values across multiple images. However, when taking images with auto exposure video cameras, this assumption typically does not hold. The automatic adjustment of the exposure times, the photometric falloff of the pixel intensities to the sides of the image due to vignetting as well as an often nonlinear camera response function cause the observed pixel intensities to differ for the same scene point. It has been shown that prior photometric camera calibration can significantly enhance the performance of DSO.

If the exposure of the camera can be controlled manually, the photometric calibration can be obtained by acquiring multiple images taken under different exposures \cite{debevec_malik_calib}\cite{mitsunaga_nayar_calib} and then estimating for a vignetting map by taking images of a uniformly colored surface \cite{dso_dataset}\cite{vignette_calibration}. However, for many video cameras the exposure times are automatically chosen and cannot be influenced by the user. Furthermore, one might want to run a visual odometry or SLAM algorithm on datasets where no photometric calibration is provided and no access to the camera is given. In these cases, it is necessary to use an algorithm that can provide calibrations for arbitrary video sequences. 

In the past a number of photometric calibration methods in the setting of an unknown camera have been proposed. Lin et al. \cite{single_image_res} introduce a single image estimation method aiming at the recovery of the response function from irradiance mixtures around edge regions using only a single input frame. Zheng et al. \cite{single_image_vig} on the other hand recover a vignetting function by identifying surfaces with identical scene radiance, measuring their photometric falloff towards the image borders. However, both of these methods can only estimate for one of these two photometric parameters, requiring knowledge of the other. Furthermore, using only a single image does not allow for estimating exposure times. 

Multiple image approaches focus on offline applications such as panorama stitching of only a few input images \cite{kim_decouple}\cite{linear_vig_estimation} for which the runtime of the algorithm is not critical and a large number of pixel correspondences can be acquired easily by aligning image pairs. Those approaches are not well suited for providing an online calibration of videos that can exhibit arbitrary motion. Nevertheless, we can adapt their underlying optimization strategies for the photometric parameter optimization. 

\begin{figure*}
	\vspace{0.2cm}
	\center
	\includegraphics[width=0.8\textwidth]{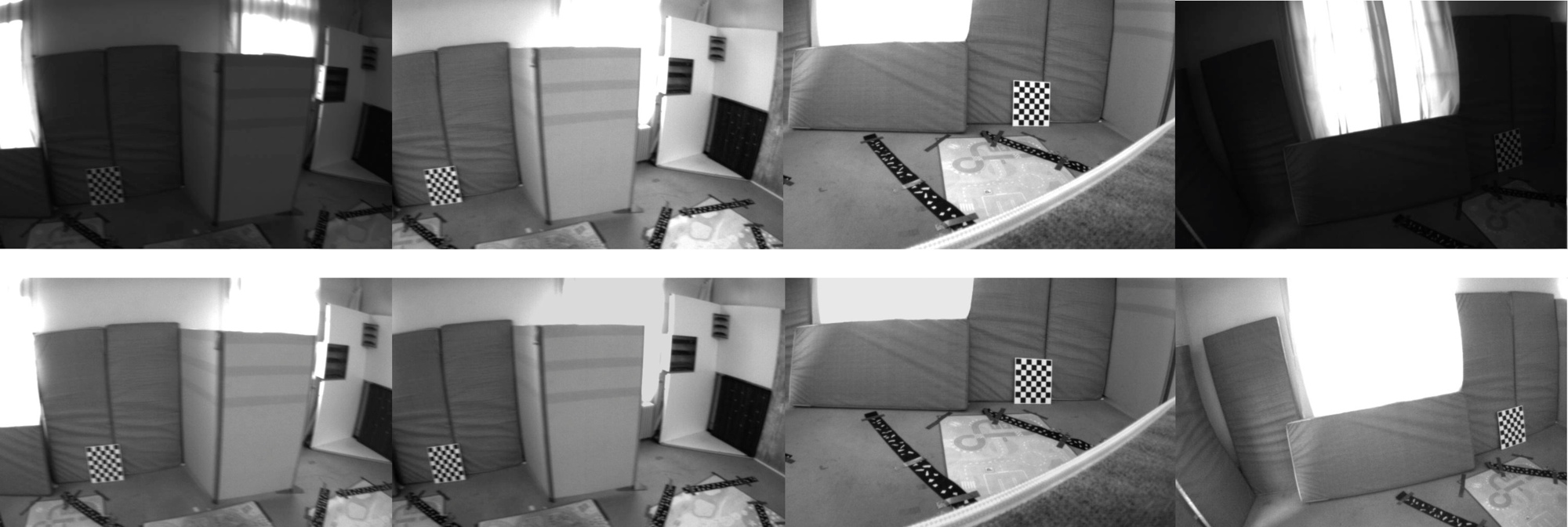}
	\caption{In this paper we propose an approach for full photometric calibration of auto exposure videos, recovering the relative exposure times, the camera response function and vignetting function, which can be used either offline for calibrating existing datasets, or online in combination with state-of-the-art direct visual odometry or SLAM pipelines. Top: sequence from the EuRoc Mav dataset with strong exposure changes. Bottom: the same sequence after photometric calibration.}
	\label{fig:euroc_qual_results}
	\vspace{-0.3cm}
\end{figure*}

Litvinov and Schechner \cite{linear_vig_estimation}\cite{linear_vig_estimation_2} propose a linear optimization framework solving jointly for response and vignetting functions in an inverse logarithmic formulation. Although their approach is computationally rather inexpensive, a linear least squares solution will be highly susceptible to outliers. Furthermore, this formulation introduces a trivial solution which must be manually excluded by introducing a model prior. To avoid these problems, Goldman et al. \cite{goldman} model the optimization problem as a nonlinear energy function which can be optimized using the Gauss-Newton algorithm. They provide a framework optimizing jointly for response function, vignetting, exposure ratios and scene point radiances. However, they also focus on panorama stitching applications, assuming that a large number of accurate correspondences can be obtained from stitching only a few images, rendering this approach impractical for auto exposure video. Furthermore, they output the joint estimate of all photometric parameters after several rounds of optimization, which will be too slow for providing an online calibration in realtime.
   
In order to avoid the necessity of performing image alignment, Kim et al. \cite{kim_tracking} proposed to use short feature tracks for calibrating arbitrary video sequences, assuming that the tracks are not affected by vignetting due to their short motion. However, calibration of the response function can only be successful in their case if heavy exposure changes are present within the tracked sequence. Also, their approach can not recover the vignetting. To allow for calibration under less drastic exposure changes, Grundmann et al. \cite{post_calib_nonlinear} propose to use long feature tracks, however also under the additional assumption that no vignetting is present.

Our algorithm builds on the work of \cite{goldman}, applying their nonlinear estimation formulation to arbitrary video sequences using gain robust feature tracking, recovering response function, vignetting, exposure times and radiances of the tracked scene points. We track features with large radial motion across multiple frames in order to recover the vignetting reliably. In the case of vignetted video, we do not require any exposure change to calibrate for the parameters, in contrast to methods which only estimate for a response function. We verify the effectiveness and accuracy of our algorithm by recovering the photometric parameters of the TUM Mono VO dataset \cite{dso_dataset} where full calibration ground truth is available as well as on manually disturbed artificial sequences of the ICL-NUIM dataset \cite{icl_dataset}. Furthermore, we show that using our algorithm in parallel to a visual odometry or visual SLAM method can significantly enhance its performance when running on datasets with photometric disturbances. Our method can also be used to improve the results of other 
methods in computer vision that rely on the brightness constancy assumption, such as for example many implementations for the optical flow problem \cite{lukas_kanade_flow}.

\section{Photometric Image Formation Process}


A scene point is illuminated by a light source and reflects the light back into space. The amount of light reflected is called the radiance $L$ of the scene point. If the radiance received by a moving observer is independent of the observers viewing angle, the scene point is called to exhibit Lambertian reflectance behavior. The radiance of the scene point is 
captured by a sensor element of the camera. The total amount of energy received at sensor location $x$ is called the Irradiance $I(x)$. 

One could expect scene points with identical radiance at different spatial locations to result in identical sensor Irradiance. 
However, for most cameras a radiometric fall off of the pixel intensities can be observed towards the image borders. This so called vignetting effect can be either due to a partial blocking of light rays by the lens barrel 
or induced by the lens geometry, modeled by the Cosine-Fourth law \cite{kim_decouple}. The irradiance $I(x)$ can therefore be obtained by multiplying the scene points radiance with a vignetting factor $V: \Omega \rightarrow [0,1]$ which is dependent on the spatial location $x \in \Omega$ of the image sensor

\vspace{-0.2cm}
$$I(x) = V(x) L.$$

When taking an image, the sensor irradiance is integrated over a time window specified by the cameras exposure time~$e$. We assume the irradiance of a sensor element to be constant over this window. The accumulated irradiance value is therefore given as $I_{acc}(x) = e I(x)$.

$I_{acc}(x)$ is then mapped by the camera response function (CRF) $f: \mathbb{R} \rightarrow [0,255]$ to an image output intensity. For real cameras, the input of the CRF is limited by the cameras dynamic range. If the accumulated irradiance falls outside of the cameras dynamic range, the scene point is under or overexposed and will be given either the pixel value $0$ or $255$ respectively. For our work, since the physical scale of the radiances cannot be recovered, the dynamic range of the camera is normalized to the unit interval $[0,1]$. 

The entire image formation process mapping a scene points radiance $L$ to an image output intensity $O$ can be compactly written as 

\vspace{-0.2cm}
\begin{equation} O = f\big(e V(x) L\big) \label{eq:main_eq}. \end{equation}

\section{Tracking Frontend}

\begin{figure}
	\vspace{0.2cm}
    \centering
    \begin{subfigure}[b]{0.49\textwidth}
        \includegraphics[width=\textwidth]{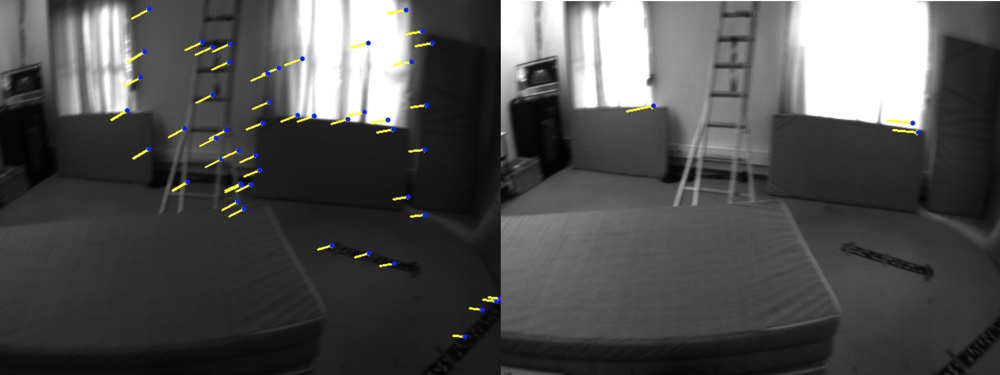}
        \label{fig:matching_raw}
    \end{subfigure}
    
    \vspace{-0.2cm}
    \begin{subfigure}[b]{0.49\textwidth}
        \includegraphics[width=\textwidth]{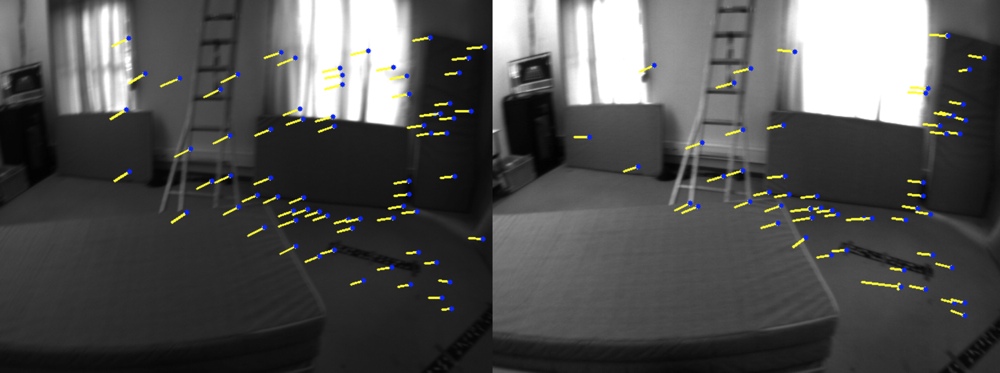}
        \label{fig:matching_raw}
    \end{subfigure}

    \vspace{-0.2cm}
    \caption{KLT tracking on a EuRoC sequence with challenging illumination. Top row: Standard KLT tracks between two images Bottom row: Gain adaptive KLT tracks between the same images. Note how the standard implementation
    of the KLT tracker cannot handle larger exposure time changes between images, while the adaptive version handles these situations well.}
    \label{fig:adaptive_tracking_benefit}
    \vspace{-0.6cm}

\end{figure}

Given a set of image frames $F$ for calibration, all the variables in Eq. \eqref{eq:main_eq} are a priori unknown except for the output intensities $O$. In order to obtain a photometric calibration, a set of scene points $P$ must be selected and their projections onto the images where they are visible must be estimated. Earlier works have used image alignment techniques to obtain a large number of pixel correspondences. However, these were mostly applied in systems performing panorama stitching and will be inaccurate in case of unconstrained camera motion. Instead we adopt a similar approach to \cite{post_calib_nonlinear}, using a pyramidal implementation of the KLT tracker to obtain point correspondences. Whereas  \cite{post_calib_nonlinear} assumes that the brightness changes are small between consecutive frames, we have observed that some datasets exhibit strong exposure changes between images and, therefore, the standard KLT tracker will fail in these cases. Hence, we use the implementation suggested by \cite{kim_tracking}, optimizing jointly for the tracking updates and a gain ratio between frames which can be done efficiently using the Schur complement. Fig. \ref{fig:adaptive_tracking_benefit}  shows the difference between the two trackers when applied on frames with larger exposure change. We extract Shi-Thomasi corners \cite{good_features_to_track} which well constrain the solution to the optical flow equation, constituting good candidate points for tracking.

In order to reliably recover the vignetting, it is important to sample features uniformly across the image to cover the entire radial range. Therefore the image is divided into a number of grid cells and a total sum of $N$ features is sampled from all the cells. This is also beneficial in order to cover the entire image intensity range which is required to constrain the response function well. In case features are lost due to occlusions or scene points moving outside of the image, new features are extracted from cells currently containing a lower number of features. We use long feature tracks in order to increase radial movements, which are necessary for vignetting estimation since, if the radial movements of the tracked features are small, the changes of irradiance due to spatial photometric fall-off will not be captured. 

\begin{figure}
	\vspace{0.3cm}
    \centering
    \begin{subfigure}[b]{0.23\textwidth}
        \includegraphics[width=\textwidth]{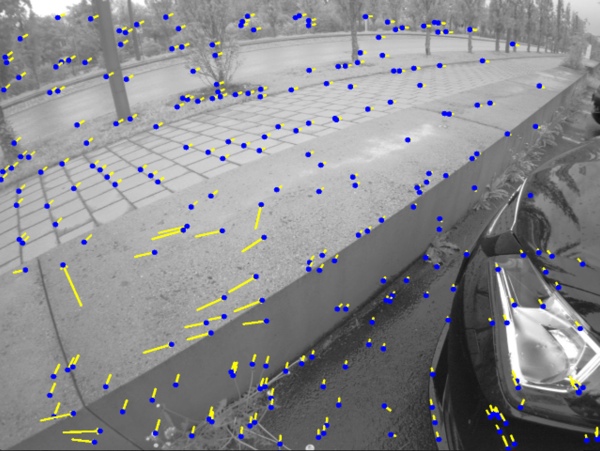}
        \label{fig:matching_raw}
    \end{subfigure}
    ~ 
    \begin{subfigure}[b]{0.23\textwidth}
        \includegraphics[width=\textwidth]{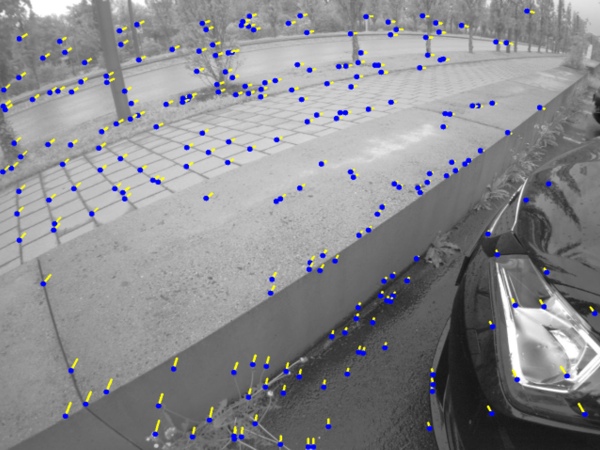}
        \label{fig:matching_rejected}
    \end{subfigure} 
    
    \caption{Comparison of tracking without and with evaluation of the forward-backward error. The right image shows tracking when filtering large forward-backward tracking inconsistencies, whilst the left image shows standard KLT tracks. Note how the spurious tracks in the left on the low gradient region are filtered out in the right image.}
    
    \label{fig:tracking_outliers}
    \vspace{-0.6cm}
\end{figure}

Since the scene points that cannot be reliably tracked over a longer time (such as low gradient regions) can heavily disturb the estimation result, the forward-backward tracking error is evaluated to early filter badly tracked correspondences \cite{bidirectional_error}. Fig. \ref{fig:tracking_outliers} shows feature tracks in one of the TUM Mono VO sequences with the error filter disabled and enabled. It can be seen that the wrong updates of the features on the low gradient region are entirely filtered out when forward-backward tracking is enabled.

In order to increase the number of tracked scene points without having to extract further features from the image, a small image patch is extracted around tracked feature locations. Another reason for why this is beneficial is that the tracked features will typically be located on large image gradients in order to be tracked reliably. Therefore, already a small geometric tracking inaccuracy will cause a large difference in image brightness for the correspondences. Extracting an image patch aims at obtaining low gradient correspondences, whose image intensity profiles will be less sensitive with respect to small tracking errors.

\section{Optimization Backend}

Given a set of scene points $P$ tracked across a range of images where point $p \in P$ is visible in frames $F_{p}$, we use Eq. \eqref{eq:main_eq} to formulate an energy as

\begin{equation} E = \sum_{p \in P} \sum_{i \in F_p} w_{i}^{p} \bigg| \bigg| \underbrace{O^{p}_{i} - f\big(e_i V(x^{p}_{i}) L^{p}\big)}_{r(f,V,e_i,L^p)}\bigg| \bigg|_{h}, \label{eq:Energy} \end{equation}

where $O^{p}_{i}$ is the output intensity of $p$ in image $i$, $e_i$ is the exposure time of image $i$, $L^{p}$ is the radiance of $p$ and $x^{p}_{i}$ is the spatial location of the projection of $p$ onto image $i$. $w_{i}^{p}$ defines a weighting factor for each residual $r$. In contrast to \cite{goldman} where a least square error metric is used, we use the Huber norm $||.||_{h}$ for robust estimation, parametrized by $h \in \mathbb{R}$. Note that this formulation of the energy assumes every scene point to be located on a Lambertian surface.

As has been noted before, there exists an exponential ambiguity in recovering the parameters when the response function is unknown  \cite{grossberg_ambiguities}. Given a solution, one can obtain a different solution yielding the same energy by choosing a constant $\gamma \in \mathbb{R}$ and defining $\tilde{f}(x) = f(x^{1/\gamma}),\tilde{V}(x) = V(x)^\gamma,\tilde{e}_i = e_i^{\gamma},\tilde{L}^{p} = ({L^{p}})^\gamma$. 
Determining $\gamma$ requires the introduction of some prior knowledge on the parameter space. In \cite{goldman} it is proposed to add a prior term to the energy pulling one point on the response function to a certain user defined value. This is especially required in the presence of a trivial solution as in $\cite{linear_vig_estimation}$, where choosing the inverse response as a constant function yields a minimum of the energy. However, since our formulation does not include such a trivial solution and a local optimization algorithm with a prior parameter initialization is used, we have not found it necessary to include any prior term to the energy. Instead, we fix the value for $\gamma$ by constraining the response to run through a certain point after optimization has finished.

To model the CRF, we use the empiric model of response (EMoR) introduced by Grossberg and Nayar \cite{grossberg_model}. A principle component analysis (PCA) is applied to find the mean response $f_0(x)$ and basis functions $h_k(x)$ which can be linearly combinated to form the overall response function $f_{G}(x)$ by choosing parameters $c_k \in \mathbb{R}$

\vspace{-0.3cm}
\begin{align*}
f_{G}(x) &= f_0(x) + \sum_{k = 1}^{n} c_k h_{k}(x) 
\end{align*}

This model has been successfully used in previous applications \cite{goldman}\cite{kim_tracking} and exhibits desirable properties for a CRF such as $f_G(0) = 0, f_G(1) = 255$ for all possible parametrizations. Furthermore its derivative can be easily obtained by simply deriving the median term and the base functions, which is necessary for calculating the Jacobians used in the nonlinear optimization. We use the first  $n = 4$ basis functions which have been shown to be sufficient to represent the empiric space of responses well \cite{kim_decouple}.

Since estimating a vignetting factor at every pixel location of the image is not feasible without a very large number of correspondences, we employ a flexible radial vignetting model as used in \cite{goldman}, assuming that the attenuation factors are symmetric around the image center. Furthermore, we assume that the center of vignetting falls together with the center of the image. It is modeled as a sixth order polynomial

$$V(x) = 1 + v_1 R(x)^2 + v_2 R(x)^4 + v_3 R(x)^6,$$

where $R(x)$ is the normalized radius of the image point $x$ with respect to the image center.

Eq. \eqref{eq:Energy} is optimized using the Gauss-Newton algorithm with Levenberg-Marquardt (LM) damping and analytic Jacobians. Since the proposed point tracking approach gives rise to a large number of irradiances to be estimated, jointly optimizing for all the parameters is computationally intensive. Therefore, the dependency structure of the optimization problem is exploited to decouple the estimation of the irradiances from the other parameters.


In the first step, the parameters for the response and vignetting function as well as the exposure times are updated by assuming the radiances of the corresponding scene points $L^p$ as constant and calculating for each residual $r$ the corresponding row of the Jacobian

$$\bm{J} = \bigg(\frac{\partial r}{\partial \bm{c}},\frac{\partial r}{\partial \bm{v}},\frac{\partial r}{\partial e_i}\bigg),$$ 

where $\bm{c} = (c_1,c_2,c_3,c_4)$ and $\bm{v} = (v_1,v_2,v_3)$. Let $\bm{e}$ denote the vector of all exposure times. The state update for the parameters $\Delta \bm{x} = (\Delta \bm{c},\Delta \bm{v},\Delta \bm{e})$ can then be found by solving the normal equation

$$ (\bm{J^T W J} + \lambda \thinspace diag(\bm{J^T W J})) \Delta \bm{x} = \bm{J^T W r}, $$

where $\bm{J}$ denotes the full Jacobian matrix, $diag(\bm{A})$ denotes the operation extracting the diagonal part of the input matrix
$\bm{A}$, $\bm{r}$ is the stacked residual vector and $\lambda \in \mathbb{R}$ is the damping factor for LM-optimization. 

The diagonal weight
matrix $\bm{W}$ is built by combining two weighting factors, one for robust Huber estimation $w_r^{(1)}$, and a second one $w_r^{(2)}$ chosen to down weight residuals at image locations of high gradient since for these points, small inaccuracies in the correspondence estimation result in a large error in image intensity. We define the gradient dependent weights $w_r^{(2)}$ for a residual $r$ as

$$ w_r = \frac{\mu}{\mu + || \nabla F_i (x^{p}_{i}) ||_{2}^{2}},$$

where $\mu \in \mathbb{R}^{+}$ is a positive constant and $|| \nabla F_i (x^{p}_{i}) ||_{2}^{2}$
is the squared L2 norm of the image gradient at
location $x^{p}_{i}$ in image $F_i$. This will typically down weight
residuals in the center of the tracked image patch whilst
giving higher importance to the residuals in the immediate
vicinity of the tracked feature point. The final weight value
for a residual is then given by multiplying the two weights
$w_r^{(1)}$ and $w_r^{(2)}$.

In the second round, the radiances of the scene points are updated. This can be done efficiently since the radiance of a scene point is independent of the radiances of other scene points and all other parameters are fixed during this optimization round. Firstly one jacobian matrix $\bm{J}^p$ is constructed for every $p \in P$ containing the partial derivatives of its residuals with respect to the scene points radiance

$$\bm{J}^p = \bigg(\frac{\partial r}{\partial L^p}\bigg).$$

The radiance update can then be computed for every scene point independently by 

$$\Delta L = \frac{\bm{J}_p^T \bm{W}_p \bm{r}_p}{(1+\lambda) \bm{J}_p^T \bm{W}_p \bm{J}_p},$$

where again $\bm{W}_p$ contains the optimization weights, $\bm{r}_p$ is the stacked residual vector and $\lambda \in \mathbb{R}$ is the damping factor of the LM iteration. 

The algorithm then performs multiple optimization rounds, alternating between the optimization of the response, vignette, and exposure times in the first step, and radiances in the second. Since every round is guaranteed to perform a downhill step with respect to the energy, this scheme will eventually converge to a local minimum. 

We distinguish between running our algorithm in an offline and an online fashion. When calibrating offline, we assume to have access to the entire input sequence and no hard constraints on the running time. In the online setting images are received frame by frame and no information about future frames can be obtained. Here, we aim at providing photometrically calibrated frames as input to a direct method running in parallel. In the following we describe the implementation differences between the two modes.

\textbf{Offline calibration.} In offline calibration mode all the photometric parameters can be optimized jointly, first tracking the entire input sequence in order to obtain scene point correspondences and afterwards optimizing Eq. \eqref{eq:main_eq} as described above. For large sequences, to speed up the estimation process, we split the input into optimization blocks of constant size and optimize for each block independently. Due to the unknown scale of the exposure times, we let the blocks overlap a certain number of frames and align the exposures using a least square metric within these overlapping regions. After a first convergence of the algorithm, we perform outlier rejection by removing a fixed percentage of the residuals that yield the largest error. Then the algorithm is run again to fit the remaining inliers better.

\textbf{Online calibration.} For an online calibration setting the photometric correction of the images must be provided as fast as possible. Therefore, it is not feasible to first collect a large number of images in order to optimize all the parameters jointly. However, in order to estimate for the vignetting and response function reliably, multiple frames are required for calibration. Since exposure times can be estimated from frame to frame, we suggest to decouple their estimation from the other parameters, providing an exposure time estimate immediately for an incoming frame. Our system keeps a state for the current vignette and response estimates, which is successively updated as more frames arrive, optimizing Eq. \eqref{eq:main_eq}. On the arrival of a new frame, it is corrected by removing the response and vignette based on the current estimate. Then we compute exposure times jointly for each of the last $M$ frames using a weighted linear least square error energy formulation which rearranges the residual term of Eq. \eqref{eq:main_eq}, removing the response and vignette from the output intensity

\vspace{-0.3cm}
\begin{equation} E = \sum_{i = 1}^{M} \sum_{p \in P_i} w_{i}^{p} \bigg( \underbrace{\frac{f^{-1}(O^{p}_{i})}{V(x^{p}_{i})} - e_i L^{p}}_{r(e_i,L^p)} \bigg)^2 \label{eq:linear_exp_est}, \end{equation}

where $P_i$ denotes the set of scene points visible in the $i$'th image and $f^{-1}$ is the inverse of the response function. Each residual is now only dependent on the exposure time of its frame and the radiance of its scene point. When fixing the radiances for the tracked points, Eq. \eqref{eq:linear_exp_est} becomes a linear optimization problem which can be solved efficiently. We initialize the radiances with the average output intensities of the tracked points. 

Since incoming frames are immediately corrected based on the current response and vignetting estimate, the rather slow joint optimization of all the photometric parameters does no longer constitute a bottleneck for performance, since it can be performed independently in the background. Since KLT tracking as well as the exposure optimization can be performed very fast, our algorithm is enabled for realtime processing.

\section{Evaluation}

\begin{figure}
\vspace{0.1cm}	
    \centering
    \begin{subfigure}[b]{0.23\textwidth}
        \includegraphics[width=\textwidth]{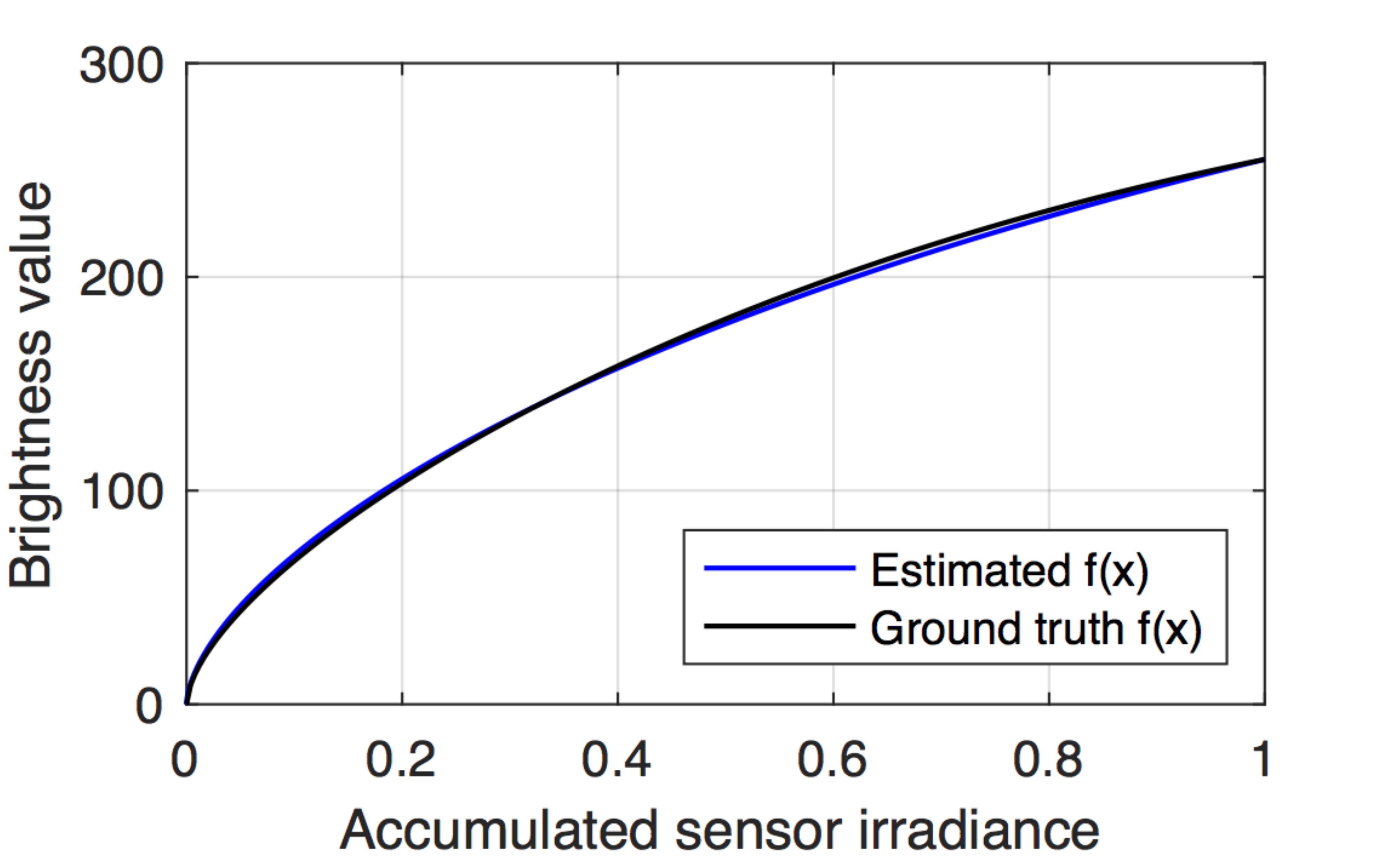}
        \caption{Response function}
        \label{fig:matching_raw}
    \end{subfigure}
    ~ 
    \begin{subfigure}[b]{0.23\textwidth}
        \includegraphics[width=\textwidth]{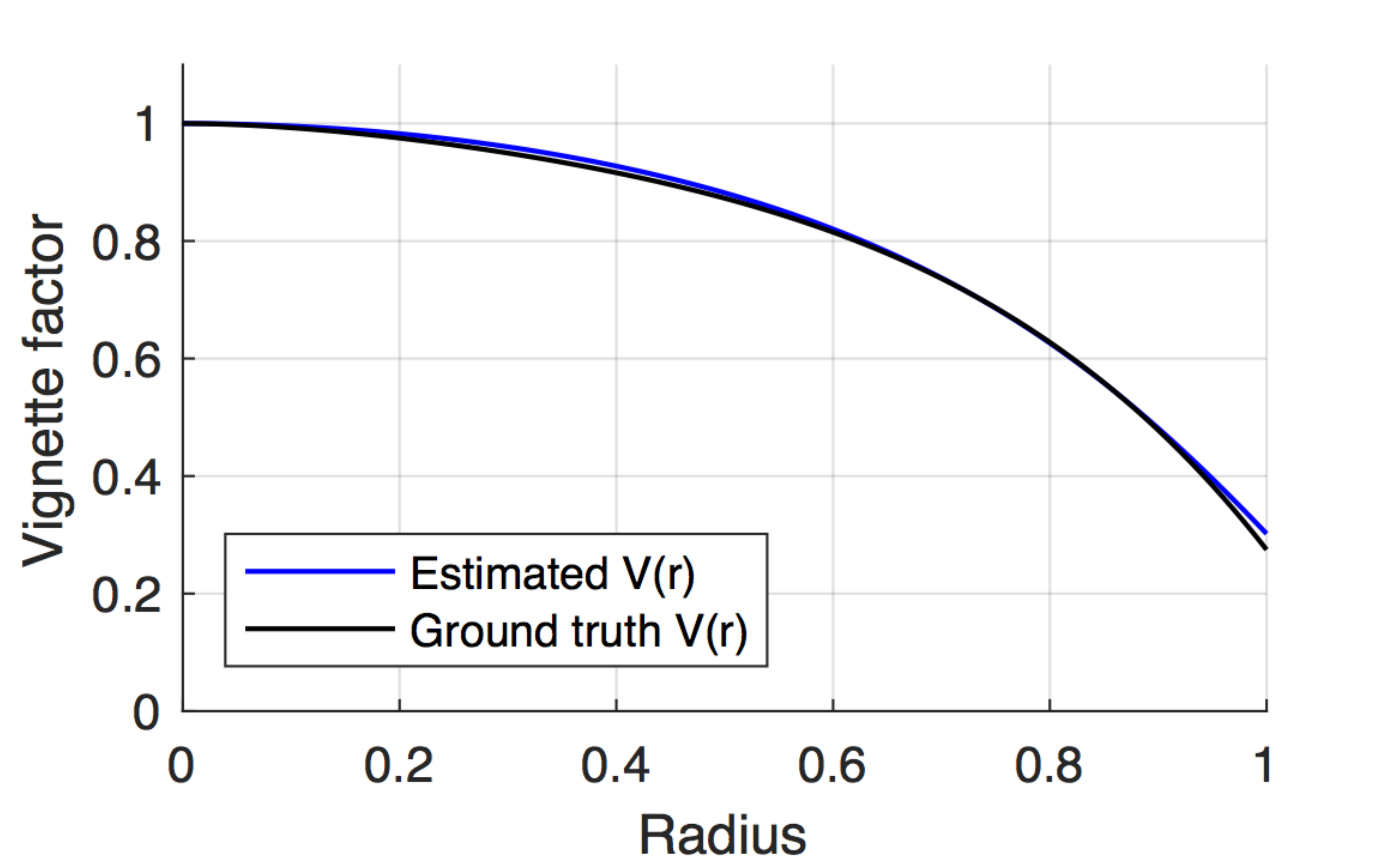}
        \caption{Vignetting}
        \label{fig:matching_rejected}
    \end{subfigure} 
    \vspace{0.2cm}
     
    \begin{subfigure}[b]{0.23\textwidth}
        \includegraphics[width=\textwidth]{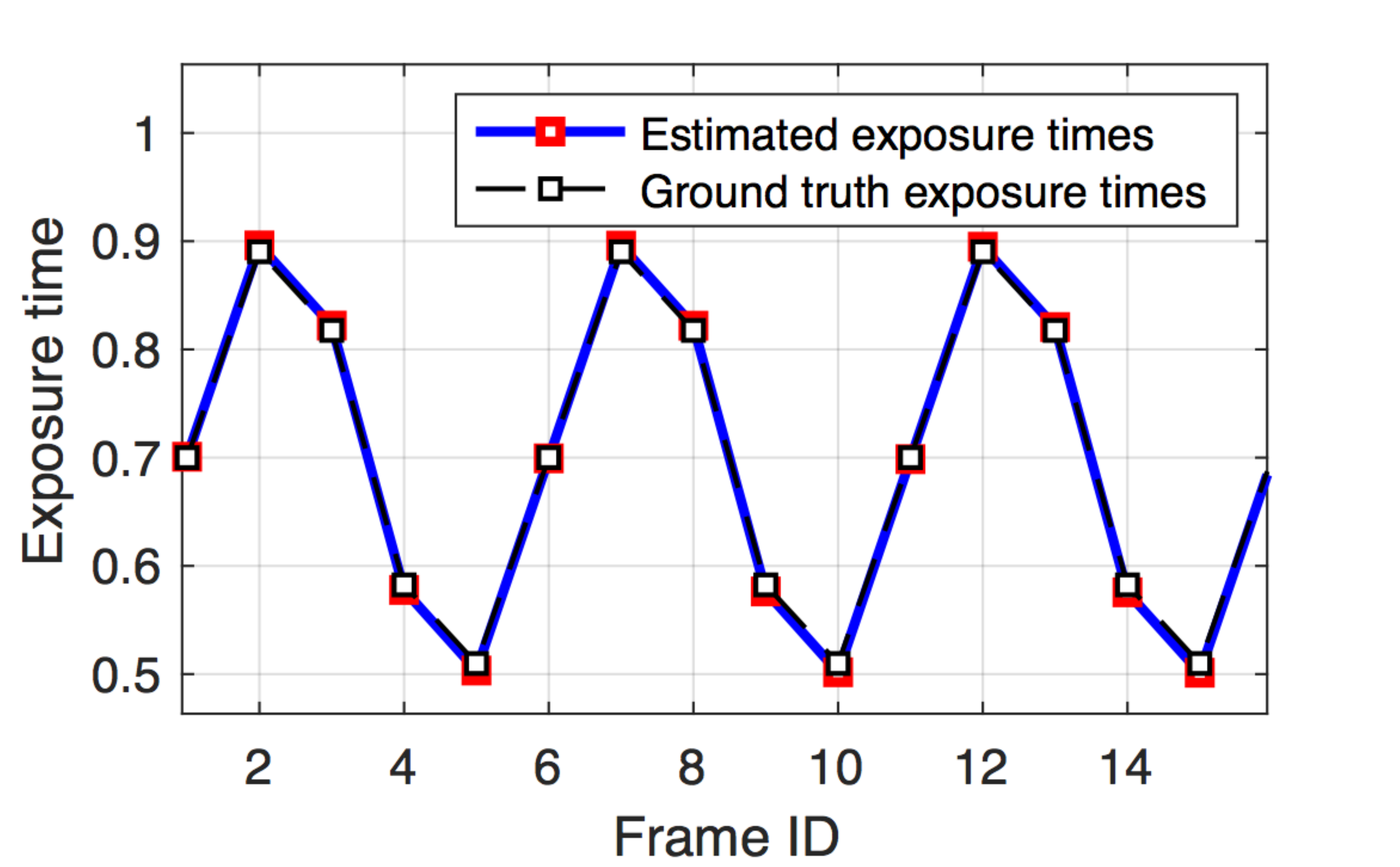}
        \caption{Exposure times}
        \label{fig:matching_rejected}
    \end{subfigure}
    ~
    \begin{subfigure}[b]{0.23\textwidth}
        \includegraphics[width=\textwidth]{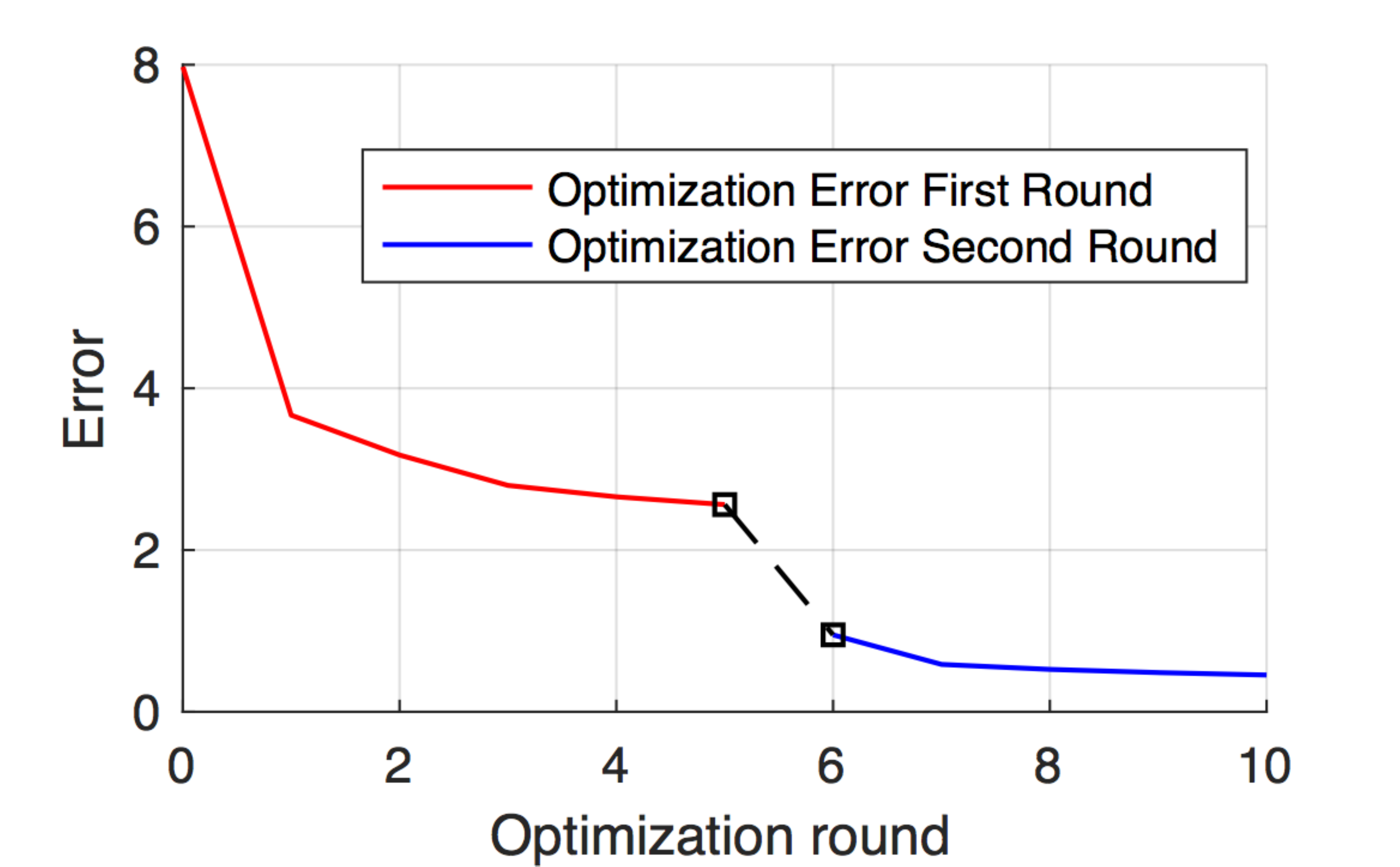}
        \caption{Errors during optimization}
        \label{fig:matching_rejected}
    \end{subfigure}
    
        \caption{Recovered photometric parameters for one sequence of the ICL-NUIM dataset with artificial photometric disturbances. Figure (d) shows the reduction of the optimization energy in every iteration before and after outlier rejection.}
    \label{fig:icl_eval}
    \vspace{-0.3cm}
\end{figure}

The implemented system is evaluated on a number of datasets, where for all sequences the tracking frontend keeps around $N = 500$ active features in every frame, extracting $5 \times 5$ image patches around each tracked feature point. The patch size is chosen as large enough to avoid only estimating using high gradient correspondences, whilst keeping the computational effort feasible. The nonlinear optimization backend is initialized with the unit response and slight vignetting. Radiances for the scene points were initially chosen as the arithmetic mean of their pixel intensities, normalized to the unit interval $[0,1]$. The grids for extracting spatially uniform features and checking for local motion consistency were constructed from cells of size $32 \times 32$ pixels.

In case of offline calibration, the input sequence is split into optimization blocks of 200 frames, where each block overlaps with its neighbors by 30 frames in order to perform the exposure time alignment. For every block the calibration is performed and the results are averaged. We find that typically the scenes will provide sufficient radial feature movement within this window for reliably calibrating for the vignetting. If the tracked scene points do not exhibit a substantial amount of radial motion across the tracked window, the estimation result is discarded since the vignetting is not well constrained in these cases. A final estimate of the vignette and response function can be obtained by averaging the results for each block. All exposure times are initially set to $1$ and optimization is performed until only an insignificant energy reduction is achieved. Then we reject $20\%$ of the residuals yielding the largest errors, performing another optimization round until convergence. 

For the online calibration setting we keep a window of the $M = 10$ most current frames in order to perform the fast exposure estimation. 
To update the response function and vignetting, the optimization backend uses blocks of $100$ tracked frames, where we only optimize for every fifth exposure time within the block. This speeds up the estimation process and is a valid approach since the frames in this block have already been photometrically corrected using the exposure time estimate from the linear optimization and the exposure times from the optimization backend are no longer required as an output. The rapidly approximated exposure times are used as an initialization for the exposure times in the nonlinear optimization. We only perform a few optimization rounds on each incoming data block without performing outlier rejection in order to further speed up our implementation.

We first evaluate our algorithm in the offline setting. The system is run on the artificially generated ICL-NUIM dataset, where photometric disturbances in response, vignetting and exposure have been applied. Figure \ref{fig:icl_eval} shows the sample recovery of the parameters, where the unknown constant for the exponential ambiguity is chosen to optimize the alignment to the ground truth with respect to a least square error metric. Since only the exposure ratios but not the absolute exposure times can be recovered (as the physical scale of the radiances is unknown), the exposure times are additionally aligned by multiplying them with a constant factor. We also show the stepwise reduction of the energy after each round, the black line indicates the energy loss after outlier rejection. 

\begin{figure}
    \centering
    \begin{subfigure}[b]{0.23\textwidth}
        \includegraphics[width=\textwidth]{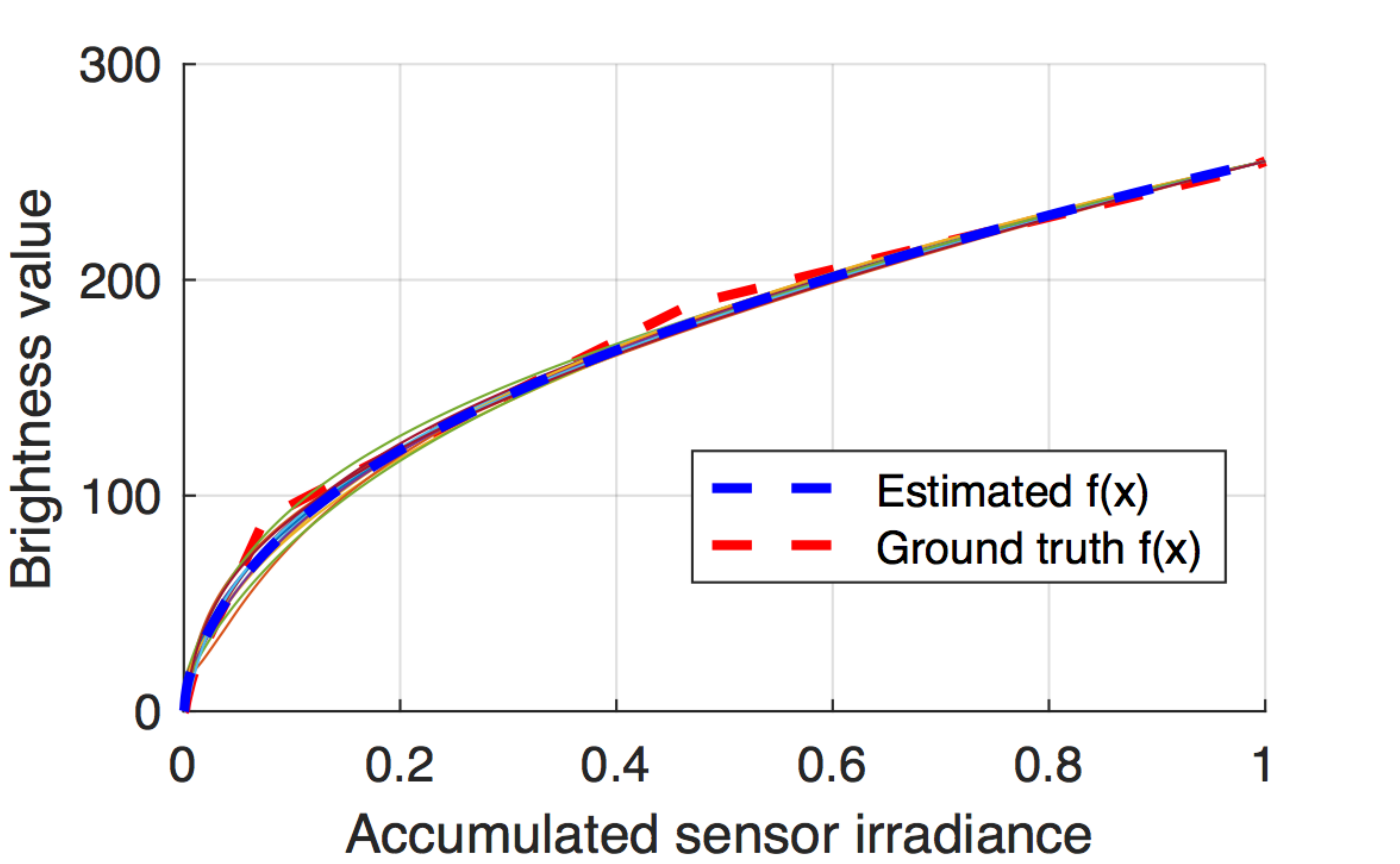}
        \caption{Response}
        \label{fig:matching_raw}
    \end{subfigure}
    ~ 
    \begin{subfigure}[b]{0.23\textwidth}
        \includegraphics[width=\textwidth]{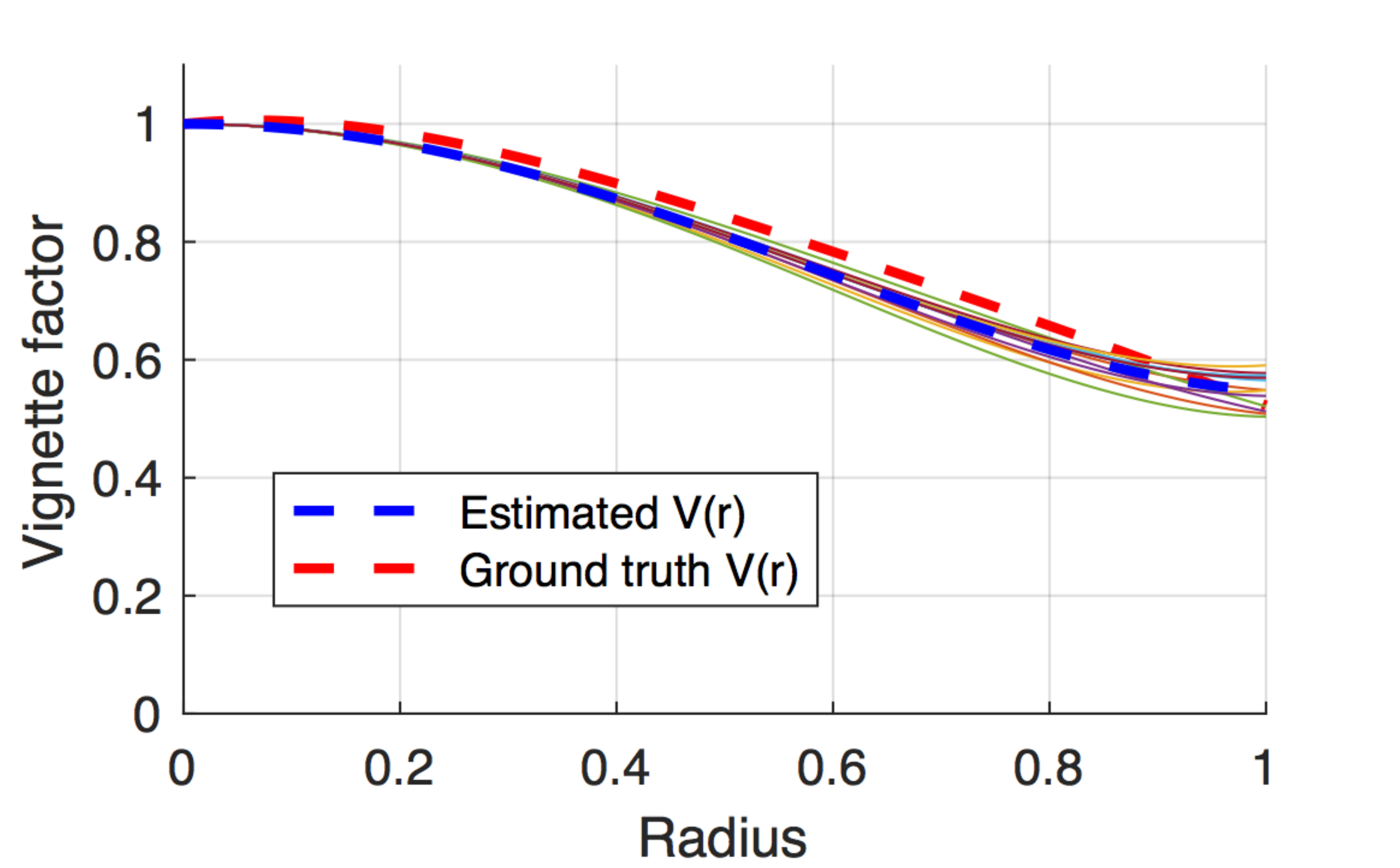}
        \caption{Vignetting}
        \label{fig:matching_rejected}
    \end{subfigure} 
     
     \vspace{0.2cm}
    \begin{subfigure}[b]{0.49\textwidth}
        \includegraphics[width=\textwidth,height=2.5cm]{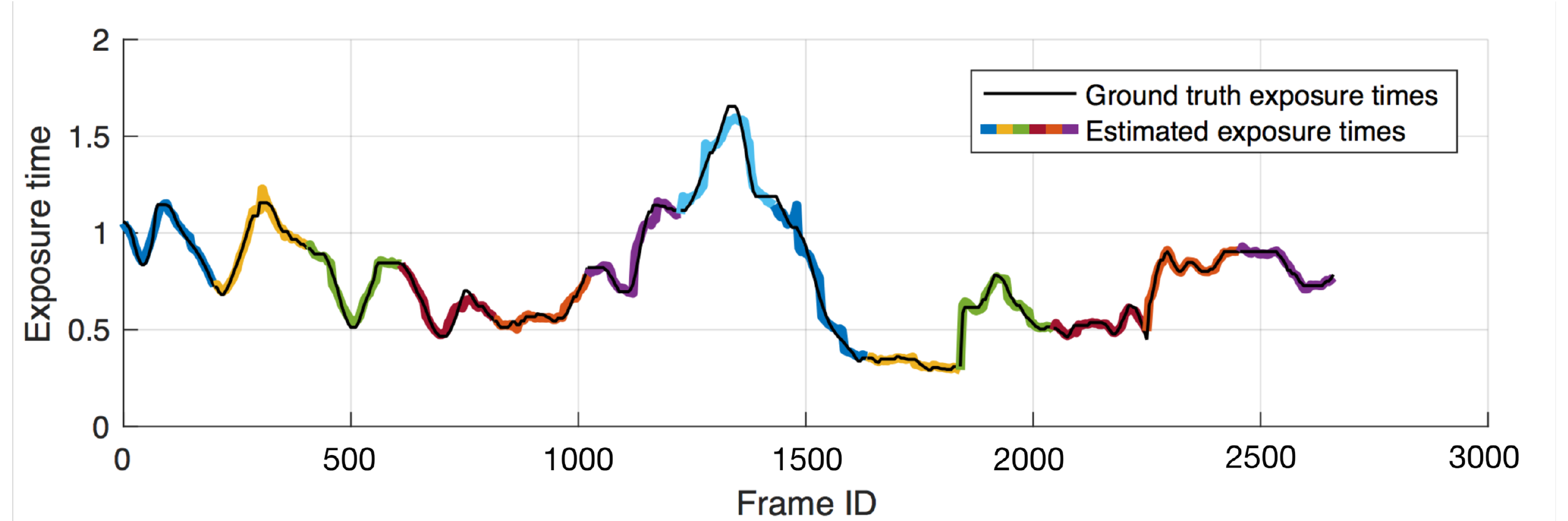}
        \caption{Exposure}
        \label{fig:matching_rejected}
    \end{subfigure}
    
        \caption{Recovered photometric parameters for sequence 50 of the TUM Mono VO dataset. Calibration was performed several times on different parts of the sequence, each time on an input block of fixed size. The recovered responses and vignetting functions as well as their averages are displayed in (a) and (b). The exposure times for each optimization block have been concatenated and are displayed in different colors in (c).}
        \vspace{-0.3cm}
    \label{fig:dso_eval}
\end{figure}

Further we ran our system on the sequences provided by the TUM Mono VO dataset for which full photometric ground truth is available. It utilizes two different cameras with different response and vignetting functions and provides ground truth for the exposure times for each frame. Fig. \ref{fig:dso_eval} shows an example offline calibration for sequence 50, showing all the calibrated vignetting functions in (a) and responses in (b), where each curve responds to one calibration block. The overall average estimation and the ground truth curves are also shown. The estimated exposure times are appended one after another and are shown in subfigure (c). The same alignment strategy as for the ICL-NUIM dataset is used. Note that for some of the optimization blocks, the exposure time changes only insignificantly, but calibration can still be performed reliably due to the presence of vignetting. 

\begin{figure}
    \centering
    \begin{subfigure}[b]{0.2\textwidth}
        \includegraphics[width=\textwidth]{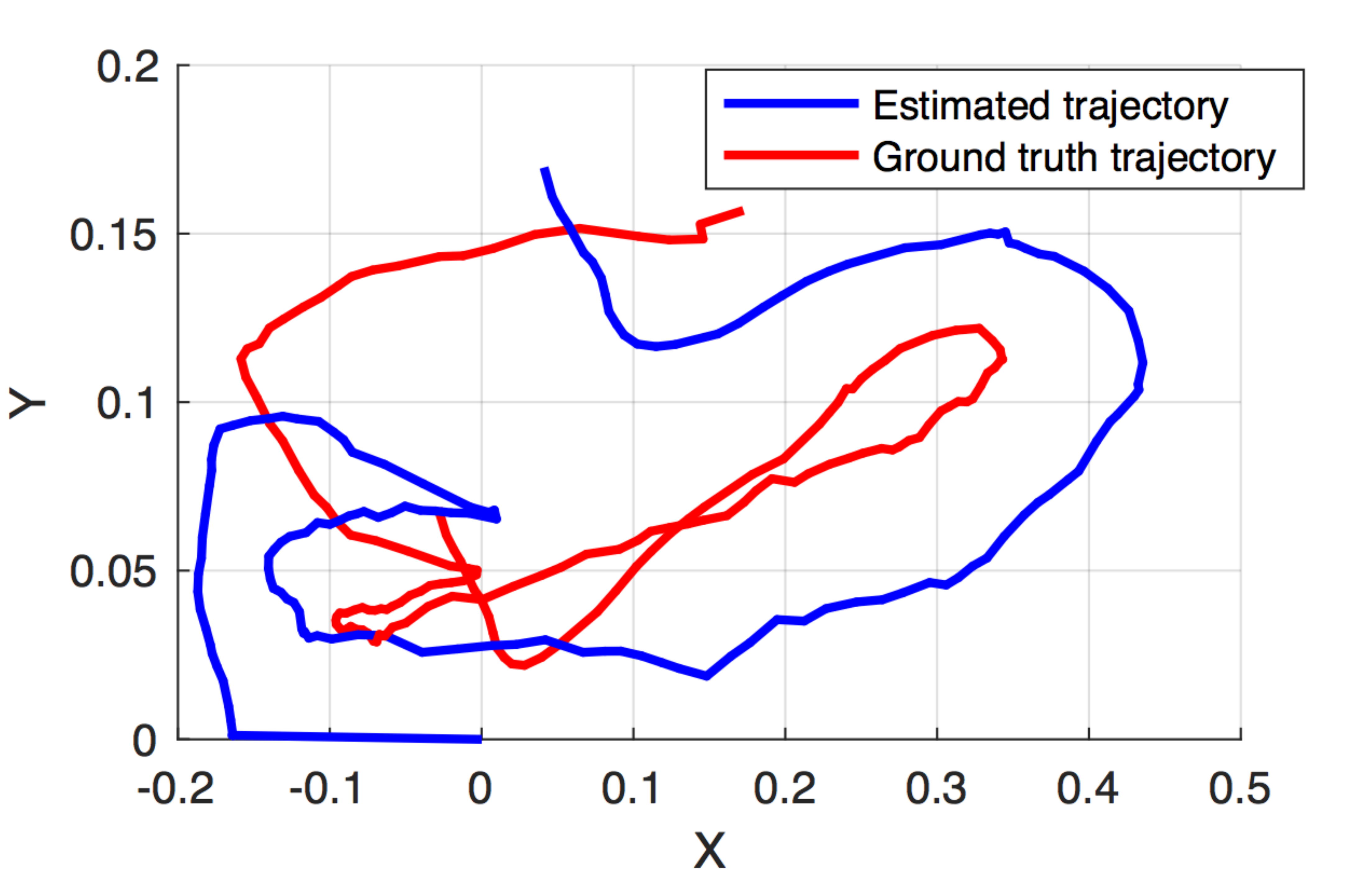}
        \caption{Without calibration}
        \label{fig:matching_raw}
    \end{subfigure}
    ~ 
    \begin{subfigure}[b]{0.2\textwidth}
        \includegraphics[width=\textwidth]{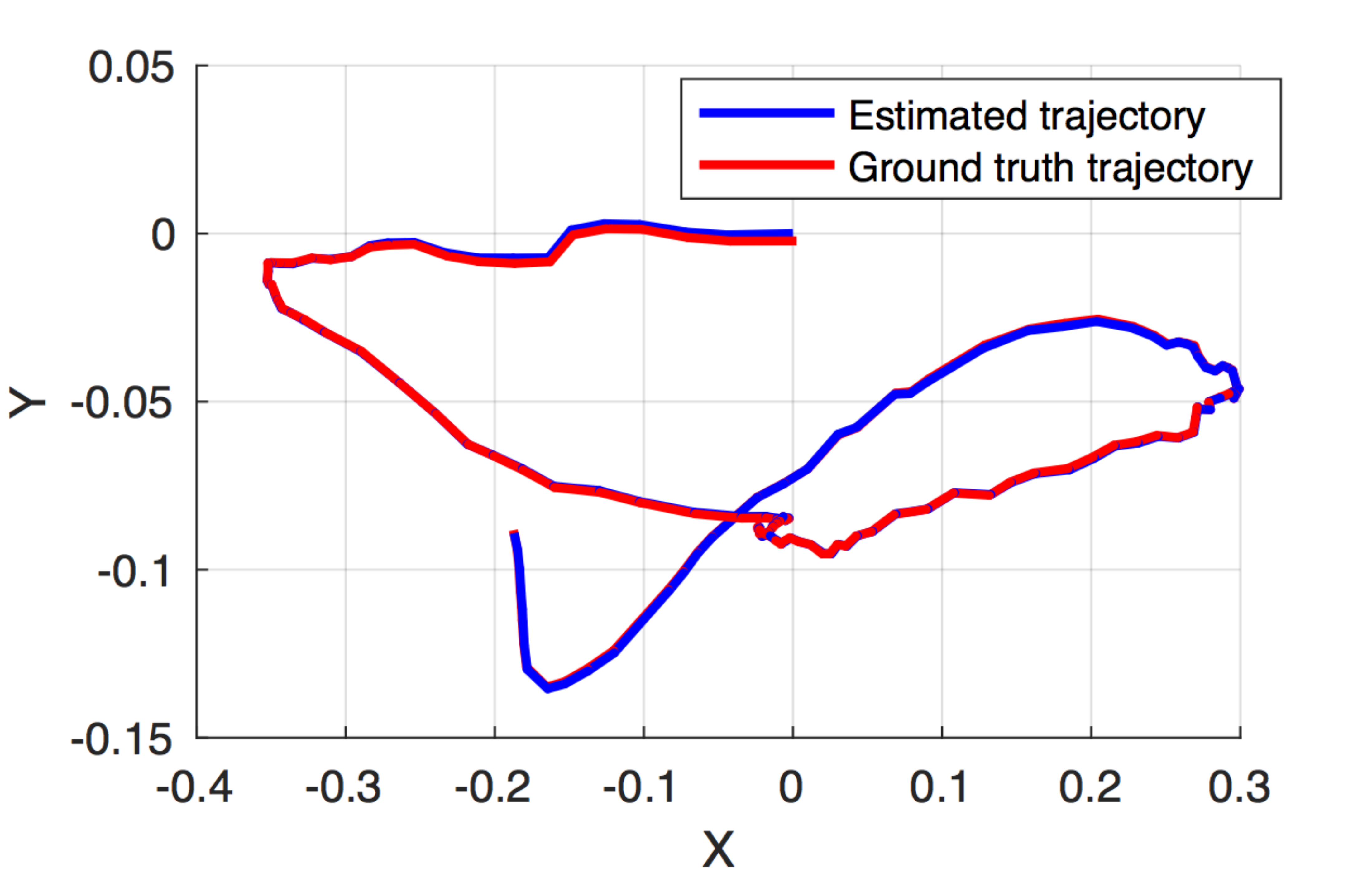}
        \caption{With calibration}
        \label{fig:matching_rejected}
    \end{subfigure} 
    
    \vspace{0.3cm}
    \begin{subfigure}[b]{0.4\textwidth}
        \includegraphics[width=\textwidth]{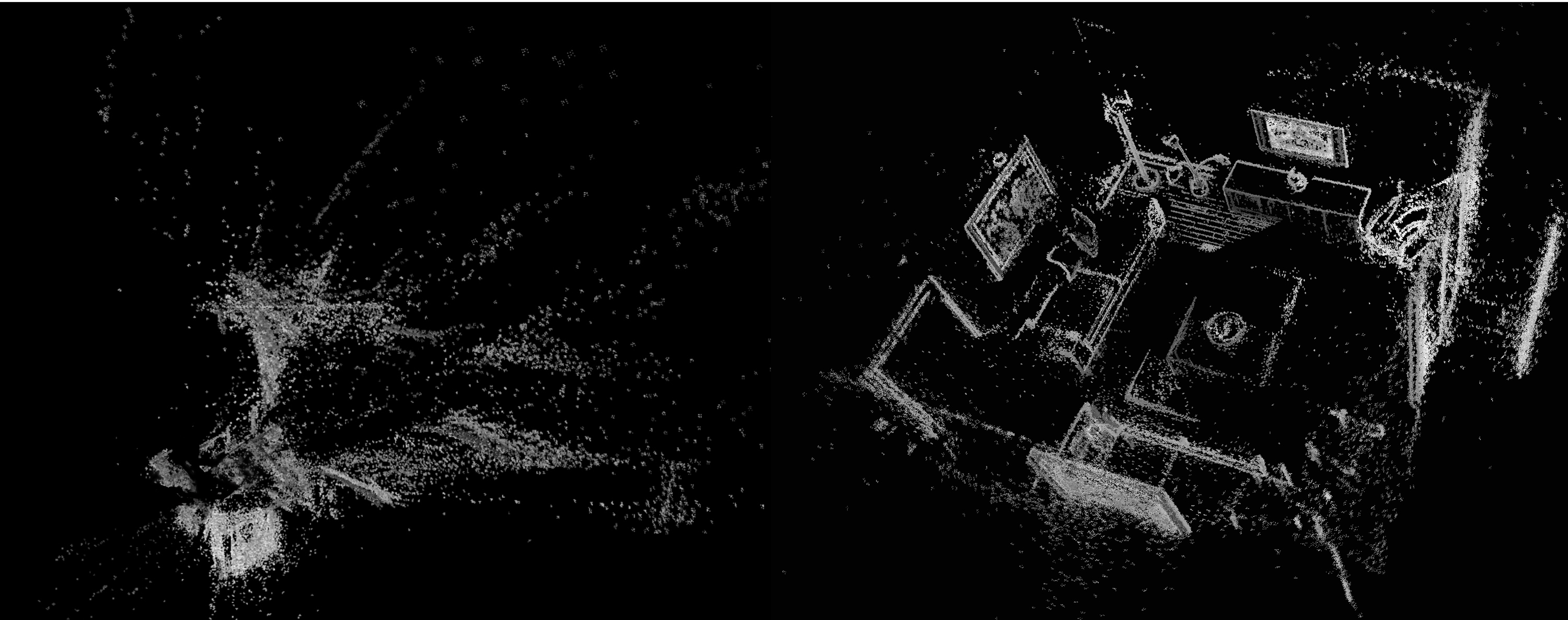}
        \caption{Reconstructed scene without and with calibration.}
        \label{fig:matching_raw}
    \end{subfigure}

    \caption{Results of running DSO on a manually disturbed sequence of the ICL-NUIM dataset with and without pre-calibration of the sequence. Note how without photometric calibration DSO entirely fails, whereas using our calibration an accurate trajectory as well as a good 3D-reconstruction are achieved.}
    \label{fig:euroc_traj_eval}
    \vspace{-0.3cm}
\end{figure}

To show the applicability of our method to visual odometry or visual SLAM, we run DSO on the 27 sequences taken with the non-fisheye camera, calibrating for the photometric parameters online. Ten runs were performed for each sequence in order to compensate for the probabilistic aspects of DSO, evaluating the alignment error as defined in \cite{dso_dataset}. Fig. \ref{fig:dso_eval} shows the alignment errors for each of the runs using either the provided ground truth, our online calibration method or no photometric calibration. Using our algorithm significantly reduces the alignment errors compared to using no calibration, yielding similar results to using the ground truth.

We also run our online version of the algorithm on a difficult sequence from the EuRoC Mav aerial dataset, which does not provide any photometric calibration data and exhibits a lot of abrupt brightness changes which are particularly challenging for direct methods to handle. Fig. \ref{fig:euroc_qual_results} shows qualitative results of the video sequence before and after correction.

\begin{figure}[!h]
    \centering
    \begin{subfigure}[b]{0.23\textwidth}
        \includegraphics[width=\textwidth]{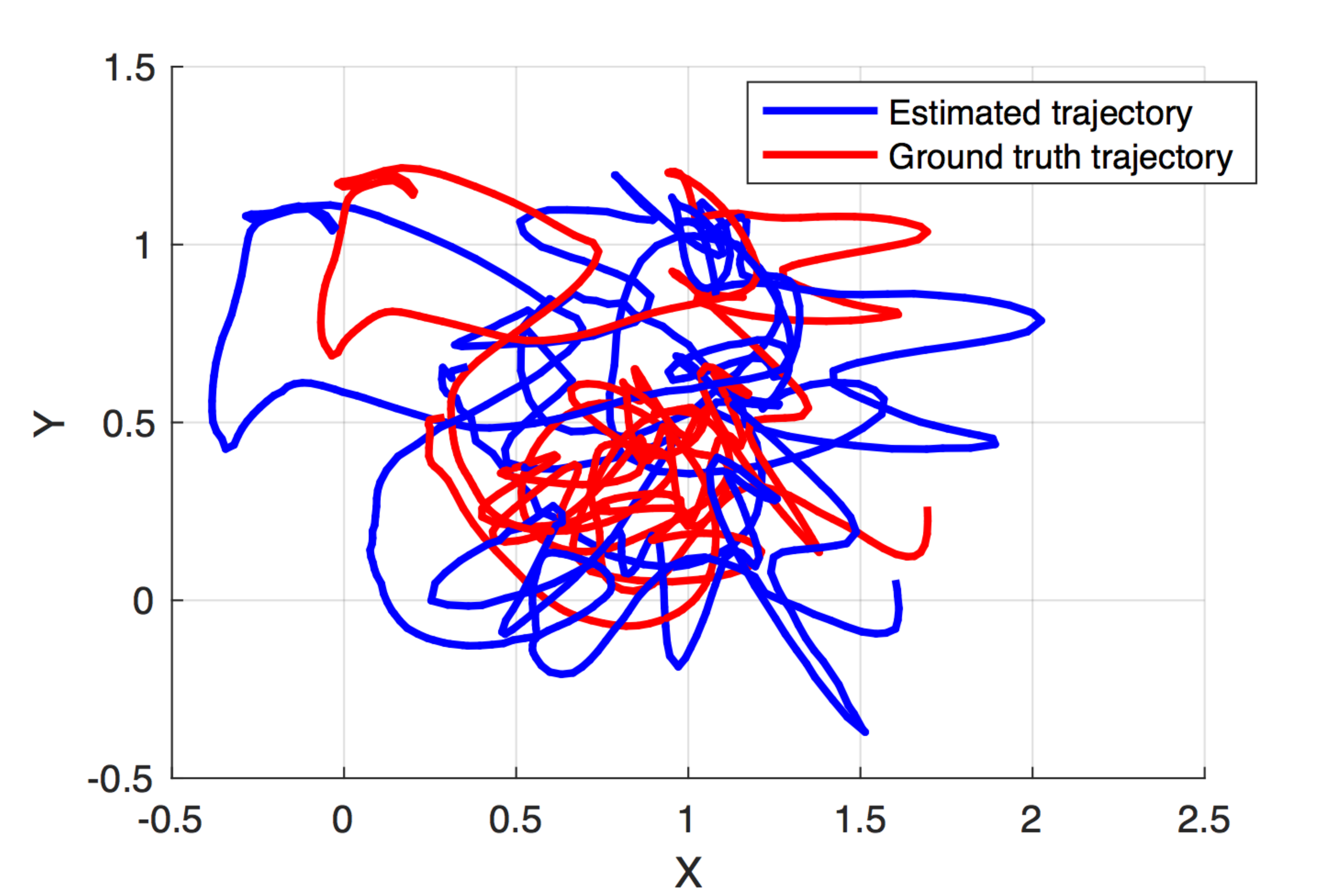}
        \caption{Without calibration}
        \label{fig:matching_raw}
    \end{subfigure}
    ~ 
    \begin{subfigure}[b]{0.23\textwidth}
        \includegraphics[width=\textwidth]{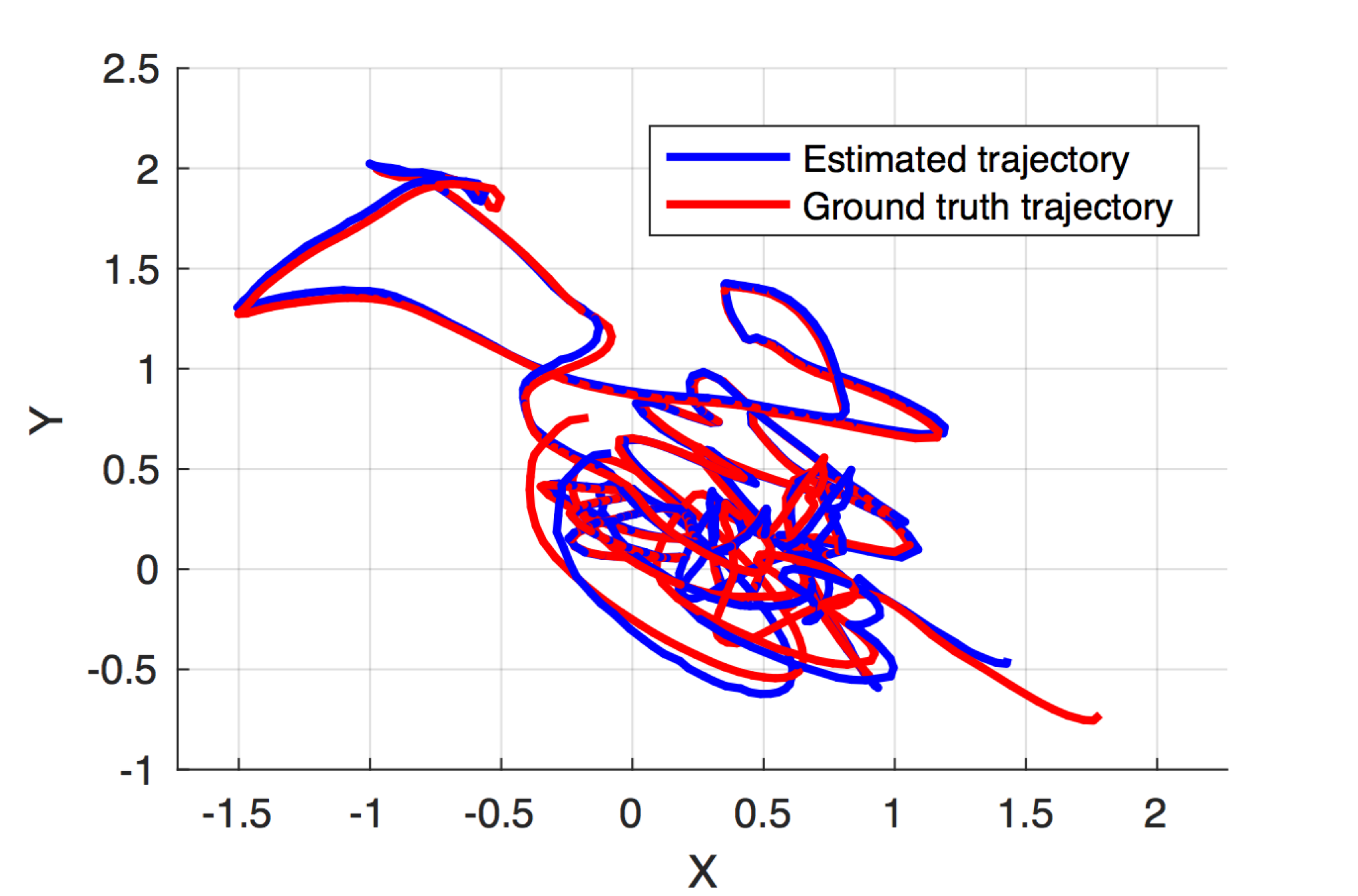}
        \caption{With calibration}
        \label{fig:matching_rejected}
    \end{subfigure} 
     
     \vspace{0.3cm}
    \begin{subfigure}[b]{0.47\textwidth}
        \includegraphics[width=\textwidth]{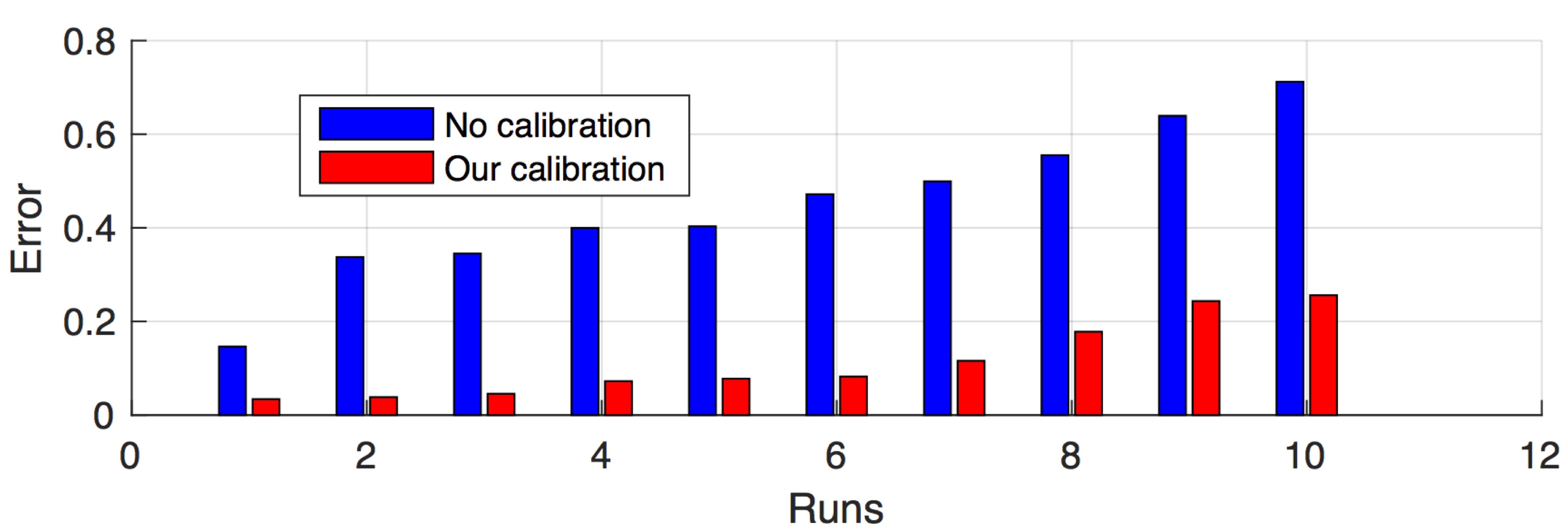}
        \caption{Trajectory error with and without calibration}
        \label{fig:matching_rejected}
    \end{subfigure}
    
    \vspace{0.3cm}
    \begin{subfigure}[b]{0.44\textwidth}
        \includegraphics[width=\textwidth]{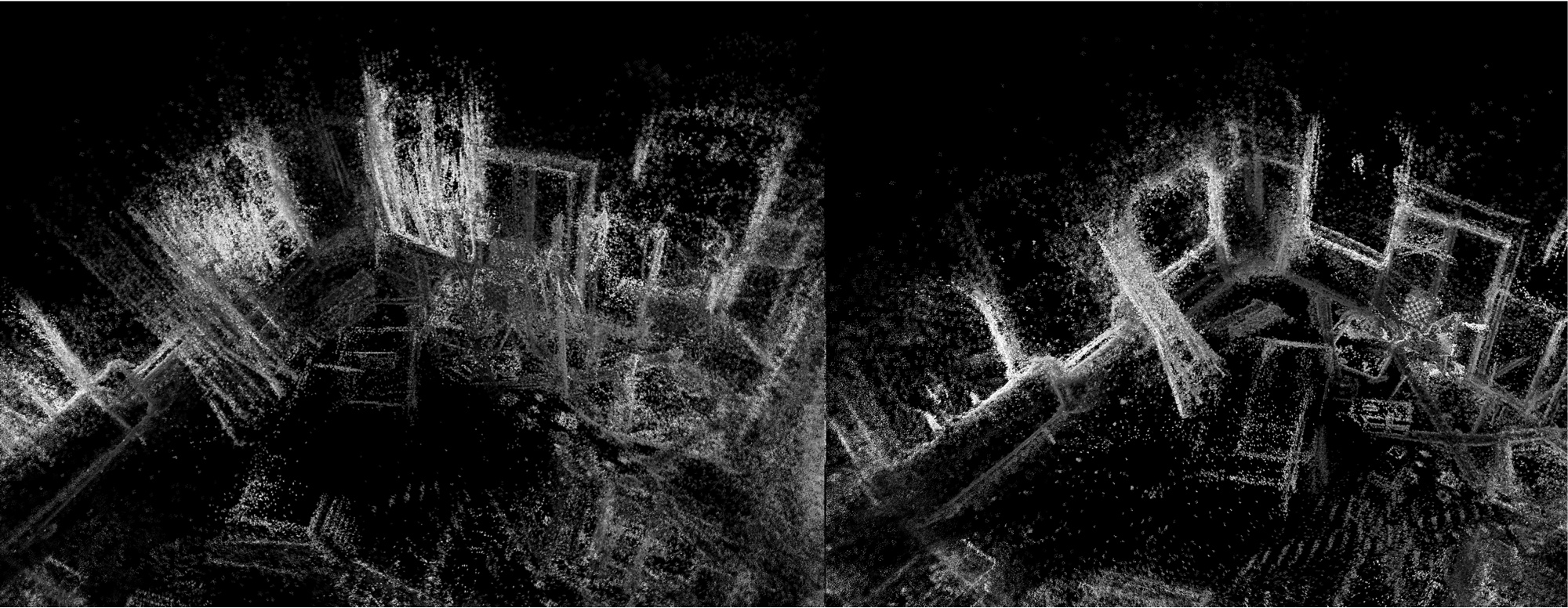}
        \caption{Qualitative result of the 3D-reconstruction without and with online photometric calibration. Note the significant improvement using our online photometric calibration as input for the direct method.}
        \label{fig:matching_rejected}
    \end{subfigure}
    
    \vspace{0.15cm}
    \caption{Results of running our algorithm on the challenging Euroc Mav sequence "Vicon room (difficult)" with and without prior photometric calibration. The trajectory error (shown for several runs of the same sequence) is significantly reduced.}
    \label{fig:euroc_traj_eval}
    \vspace{-0.3cm}
\end{figure}

Fig. \ref{fig:euroc_traj_eval} shows a trajectory obtained by running DSO on the difficult sequence of EuRoC using either no photometric calibration or using the online output of our calibrated method. Since DSO as a monocular visual odometry method does not provide the trajectory scale, a similarity transformation is calculated to align the estimated trajectory with the ground truth. Again, in order to account for probabilistic differences within the results, we run the algorithm 10 times on both input videos. Whilst in the case of no calibration, the direct method yields large trajectory errors as well as large errors in the 3D-reconstruction, we are able to achieve results with much lower trajectory error and significantly improve the reconstruct result. This also shows that the affine brightness optimization provided by DSO does not work well for drastic exposure changes.

\begin{figure}
    \centering
        \includegraphics[width=0.40\textwidth,height=3cm]{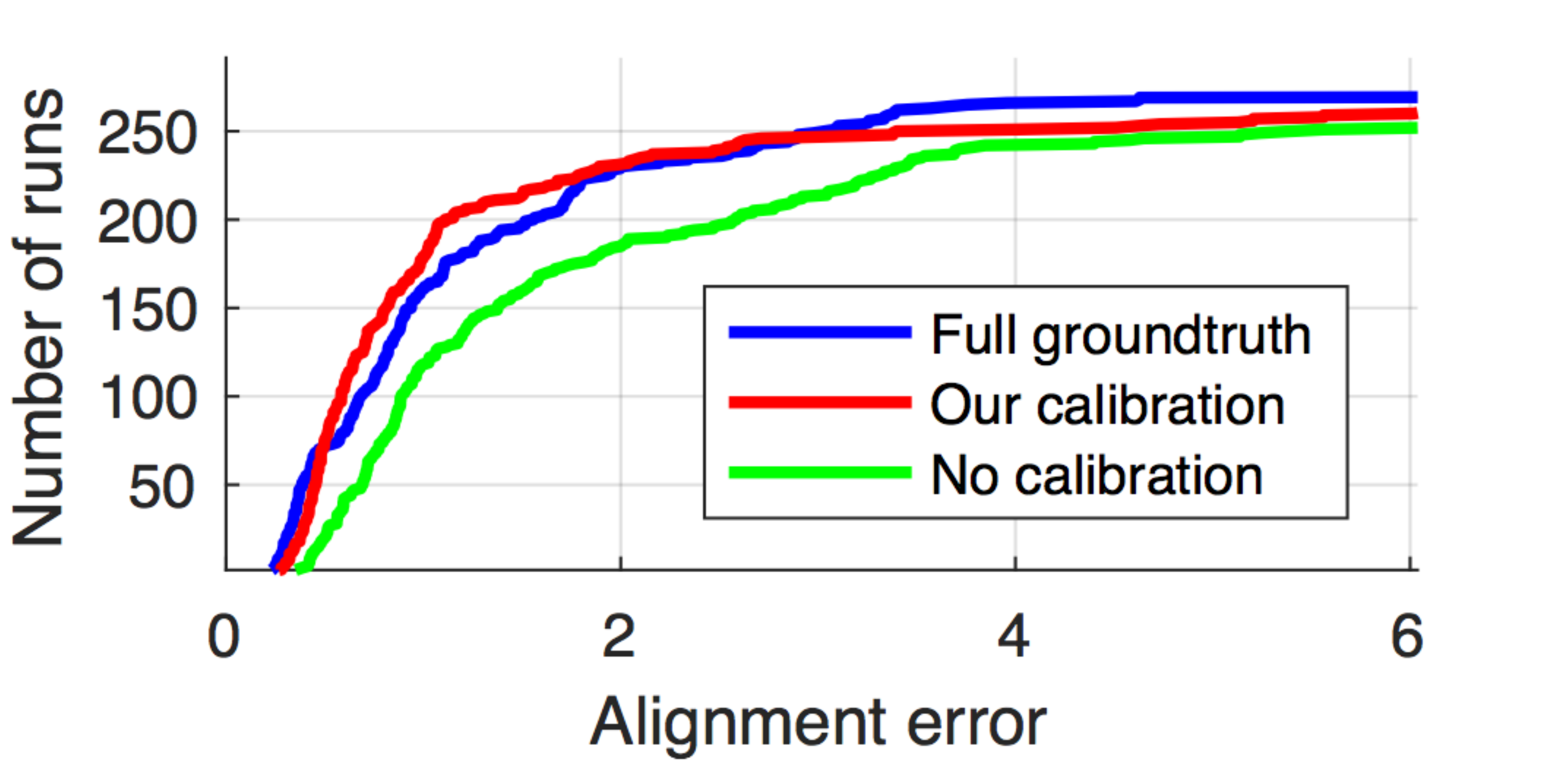}
 
    \caption{Comparison of the alignment error for running DSO on sequences for one of the cameras of the TUM Mono VO dataset using either the full ground truth calibration, no photometric calibration or our online calibration.}
    \label{fig:dso_eval}
    \vspace{-0.5cm}
\end{figure}

\vspace{-0.15cm}
\section{Conclusion}

We propose a novel system for providing realtime online calibration of auto exposure video for direct formulations of visual odometry and visual SLAM, estimating for the photometric response function, vignetting, and exposure times. We show that our algorithm provides an accurate and robust photometric calibration for arbitrary video sequences and significantly enhances the quality of direct methods for visual odometry such as DSO. In future work, we plan to substitute the KLT tracking by integrating the proposed online optimization into existing direct methods where correspondences are obtained based on a brightness constancy assumption to jointly optimize for the photometric parameters, the camera poses, and depth values.


\addtolength{\textheight}{-12cm}   









\end{document}